\documentclass{article}

\usepackage{amsmath}
\usepackage{caption}
\usepackage{subcaption}
\usepackage{multirow}
\usepackage{multicol}

\usepackage{arxiv}

\usepackage[utf8]{inputenc} 
\usepackage[T1]{fontenc}    
\usepackage{hyperref}       
\usepackage{url}            
\usepackage{booktabs}       
\usepackage{amsfonts}       
\usepackage{nicefrac}       
\usepackage{microtype}      
\usepackage{lipsum}
\usepackage{graphicx}
\graphicspath{ {./images/} }

\title{A Survey on Contrastive Self-supervised Learning}

\author{
 Ashish Jaiswal \\
  The University of Texas at Arlington\\
  Arlington, TX 76019 \\
  \texttt{ashish.jaiswal@mavs.uta.edu} \\
   \And
 Ashwin Ramesh Babu \\
  The University of Texas at Arlington\\
  Arlington, TX 76019 \\
  \texttt{ashwin.rameshbabu@mavs.uta.edu} \\
  \And
 Mohammad Zaki Zadeh \\
  The University of Texas at Arlington\\
  Arlington, TX 76019 \\
  \texttt{mohammad.zakizadehgharie@mavs.uta.edu} \\
  \And
 Debapriya Banerjee \\
  The University of Texas at Arlington\\
  Arlington, TX 76019 \\
  \texttt{debapriya.banerjee2@mavs.uta.edu} \\
  \And
 Fillia Makedon \\
  The University of Texas at Arlington\\
  Arlington, TX 76019 \\
  \texttt{makedon@uta.edu} \\
}

\begin{document}
\maketitle
\begin{abstract}
Self-supervised learning has gained popularity because of its ability to avoid the cost of annotating large-scale datasets. It is capable of adopting self-defined pseudo labels as supervision and use the learned representations for several downstream tasks. Specifically, contrastive learning has recently become a dominant component in self-supervised learning methods for computer vision, natural language processing (NLP), and other domains. It aims at embedding augmented versions of the same sample close to each other while trying to push away embeddings from different samples. This paper provides an extensive review of self-supervised methods that follow the contrastive approach. The work explains commonly used pretext tasks in a contrastive learning setup, followed by different architectures that have been proposed so far. Next, we have a performance comparison of different methods for multiple downstream tasks such as image classification, object detection, and action recognition. Finally, we conclude with the limitations of the current methods and the need for further techniques and future directions to make substantial progress. 
\end{abstract}


\section{Introduction}
    The advancements in deep learning have elevated it to become one of the core components in most intelligent systems in existence. The ability to learn rich patterns from the abundance of data available today has made deep neural networks (DNNs) a compelling approach in the majority of computer vision (CV) tasks such as image classification, object detection, image segmentation, activity recognition as well as natural language processing (NLP) tasks such as sentence classification, language models, machine translation, etc. However, the supervised approach to learning features from labeled data has almost reached its saturation due to intense labor required in manually annotating millions of data samples. This is because most of the modern computer vision systems (that are supervised) try to learn some form of image representations by finding a pattern between the data points and their respective annotations in large datasets.  Works such as GRAD-CAM \cite{selvaraju2017grad} have proposed techniques that provide visual explanations for decisions made by a model to make them more transparent and explainable. 

Traditional supervised learning approaches heavily rely on the amount of annotated training data available. Even though there's a plethora of data available out there, the lack of annotations has pushed researchers to find alternative approaches that can leverage them. This is where self-supervised methods plays a vital role in fueling the progress of deep learning without the need for expensive annotations and learn feature representations where data itself provides supervision.

\begin{figure}[ht]
\centering
    \includegraphics[width=0.65\linewidth]{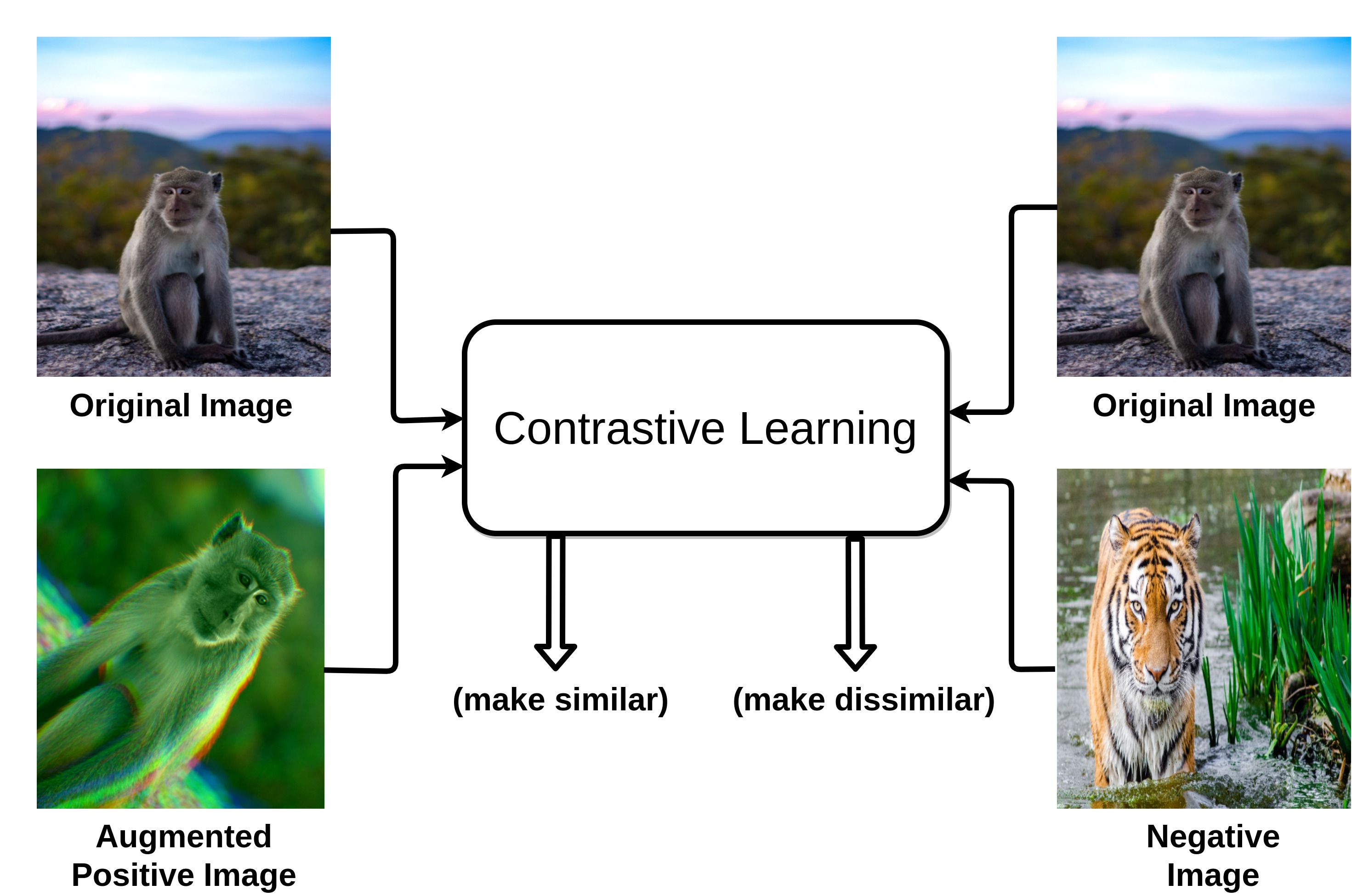}
    \caption{Basic intuition behind contrastive learning paradigm: push original and augmented images closer and push original and negative images away}
    \label{fig:cl_example}
\centering
\end{figure}

Supervised learning not only depends on expensive annotations but also suffers from issues such as generalization error, spurious correlations, and adversarial attacks \cite{liu2020self}. Recently, self-supervised learning methods have integrated both generative and contrastive approaches that have been able to utilize unlabeled data to learn the underlying representations. A popular approach has been to propose various pretext tasks that help in learning features using pseudo-labels. Tasks such as image-inpainting, colorizing greyscale images, jigsaw puzzles, super-resolution, video frame prediction, audio-visual correspondence, etc have proven to be effective for learning good representations.

\begin{figure}[hbt!]
    \includegraphics[width=\linewidth]{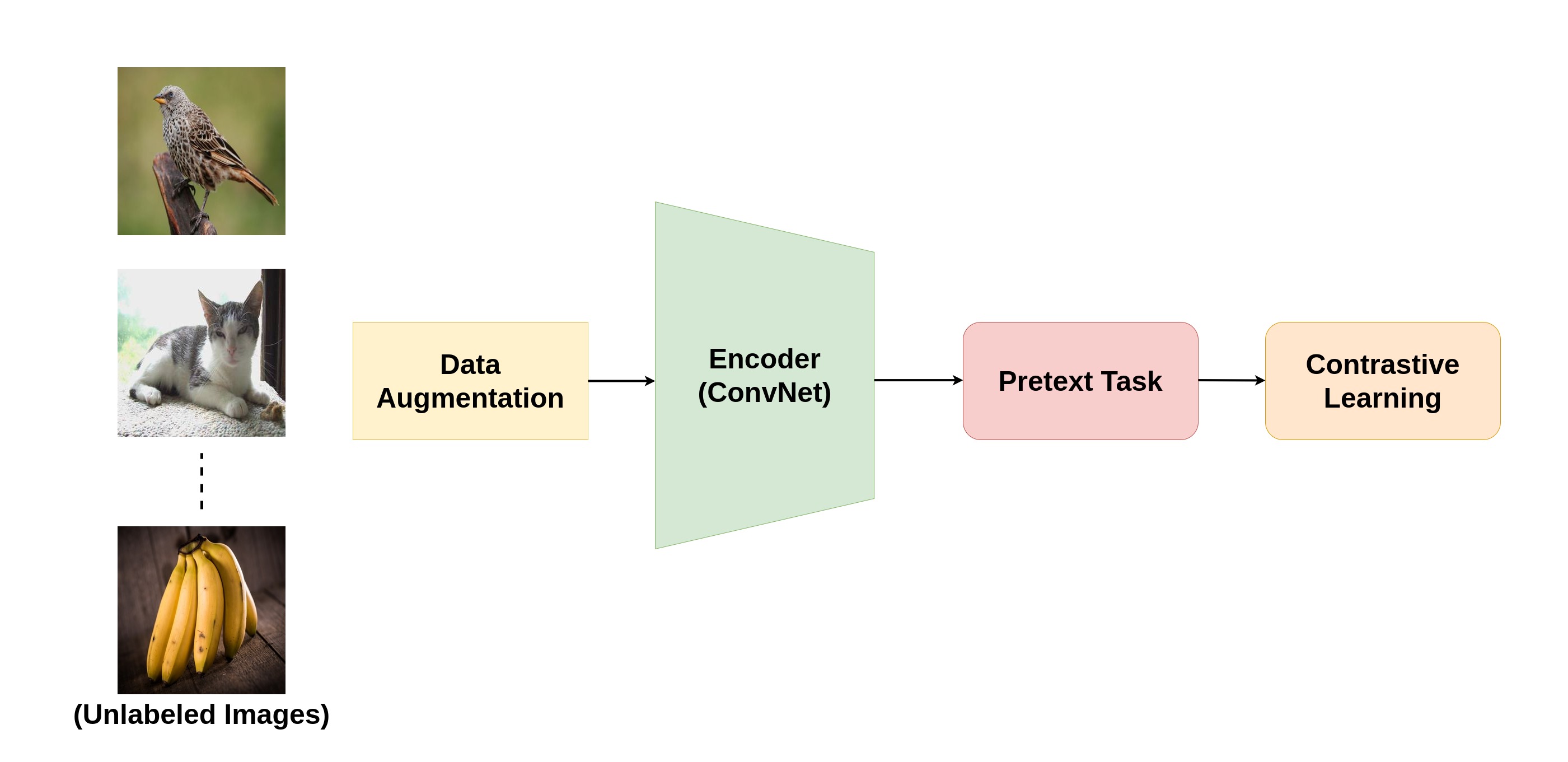}
    \caption{Contrastive learning pipeline for self-supervised training}
    \label{fig:cl_pipeline}
\end{figure}

Generative models gained its popularity after the introduction of Generative Adversarial Networks (GANs) \cite{goodfellow2014generative} in 2014. The work later became the foundation for many successful architectures such as CycleGAN \cite{zhu2017unpaired}, StyleGAN \cite{karras2019style}, PixelRNN \cite{oord2016pixel}, Text2Image \cite{reed2016generative}, DiscoGAN \cite{kim2017learning}, etc. These methods inspired more researchers to switch to training deep learning models with unlabeled data in an self-supervised setup. Despite their success, researchers started realizing some of the complications in GAN-based approaches.  They are harder to train because of two main reasons: (a) non-convergence--the model parameters oscillate a lot and rarely converge, and (b) the discriminator gets too successful that the generator network fails to create real-like fakes due to which the learning cannot be continued. Also, proper synchronization is required between the generator and the discriminator that prevents the discriminator to converge and the generator to diverge.

\begin{figure}[hbt!]
    \includegraphics[width=\linewidth]{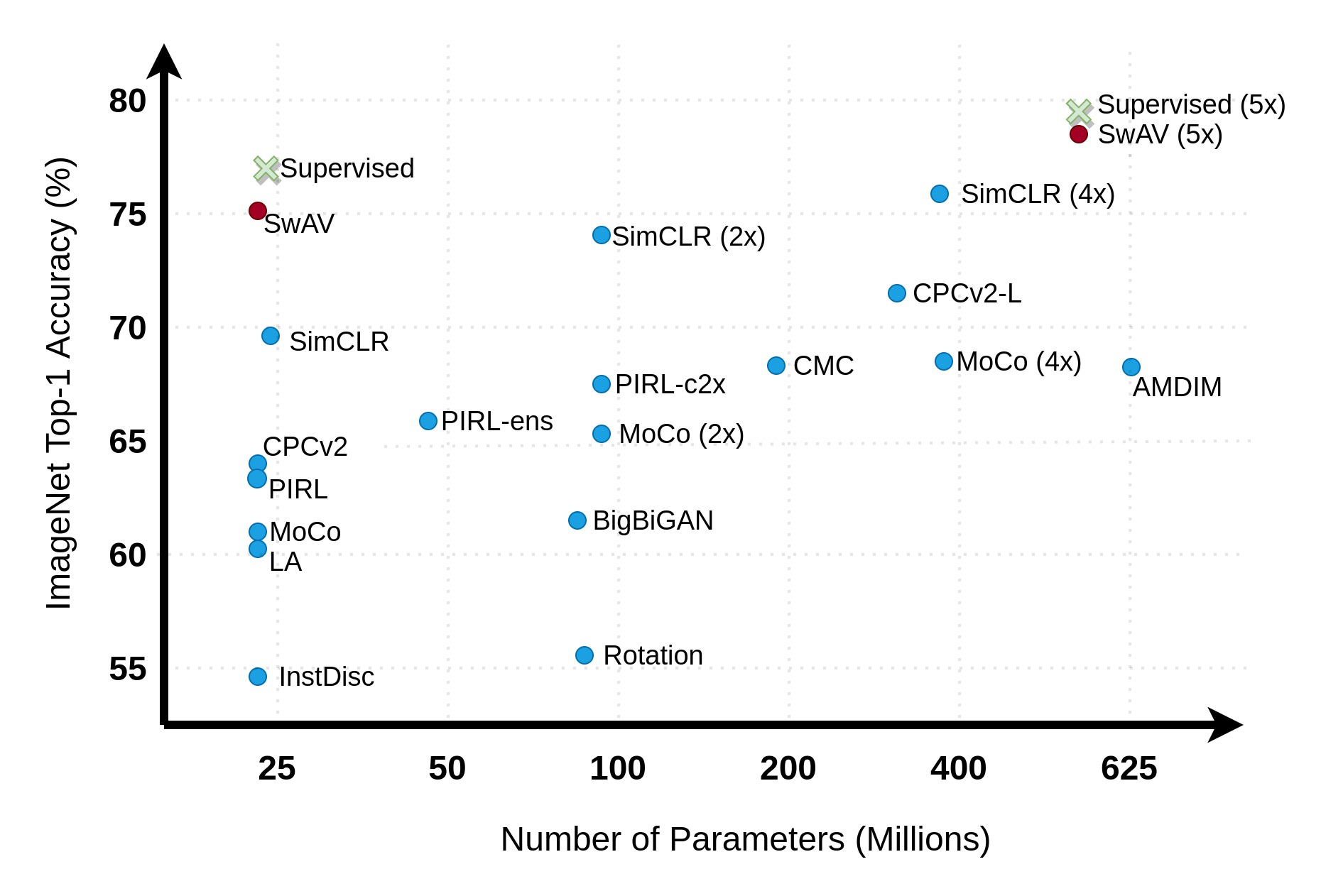}
    \caption{Top-1 classification accuracy of different contrastive learning methods against baseline supervised method on ImageNet}
    \label{fig:cl_baselines}
\end{figure}


Unlike generative models, contrastive learning (CL) is a discriminative approach that aims at grouping similar samples closer and diverse samples far from each other as shown in figure \ref{fig:cl_example}.  To achieve this, a similarity metric is used to measure how close two embeddings are. Especially, for computer vision tasks, a contrastive loss is evaluated based on the feature representations of the images extracted from an encoder network. For instance, one sample from the training dataset is taken and a transformed version of the sample is retrieved by applying appropriate data augmentation techniques. During training referring to figure \ref{fig:cl_pipeline}, the augmented version of the original sample is considered as a positive sample, and the rest of the samples in the batch/dataset (depends on the method being used) are considered negative samples. Next, the model is trained in a way that it learns to differentiate positive samples from the negative ones. The differentiation is achieved with the help of some pretext task (explained in section \ref{section:pretext}). In doing so, the model learns quality representations of the samples and is used later for transferring knowledge to downstream tasks. This idea is advocated by an interesting experiment conducted by Epstein \cite{epstein2016} in 2016, where he asked his students to draw a dollar bill with and without looking at the bill. The results from the experiment show that the brain does not require complete information of a visual piece to differentiate one object from the other. Instead, only a rough representation of an image is enough to do so.

Most of the earlier works in this area combined some form of instance-level classification approach\cite{bojanowski2017unsupervised}\cite{dosovitskiy2014discriminative}\cite{wu2018unsupervised} with contrastive learning and were successful to some extent. However, recent methods such as SwAV \cite{caron2020unsupervised}, MoCo \cite{he2019momentum}, and SimCLR \cite{chen2020simple} with modified approaches have produced results comparable to the state-of-the-art supervised method on ImageNet \cite{deng2009imagenet} dataset as shown in figure \ref{fig:cl_baselines}. Similarly, PIRL \cite{misra2019selfsupervised}, Selfie \cite{trinh2019selfie}, and \cite{tian2020makes} are some papers that reflect the effectiveness of the pretext tasks being used and how they boost the performance of their models.


\section{Pretext Tasks} \label{section:pretext}
    Pretext tasks are self-supervised tasks that act as an important strategy to learn representations of the data using pseudo labels. These pseudo labels are generated automatically based on the attributes found in the data.  The learned model from the pretext task can be used for any downstream tasks such as classification, segmentation, detection, etc. in computer vision. Furthermore, these tasks can be applied to any kind of data such as image, video, speech, signals, and so on.  For a pretext task in contrastive learning, the original image acts as an anchor, its augmented(transformed) version acts as a positive sample, and the rest of the images in the batch or in the training data act as negative samples.

Most of the commonly used pretext tasks are divided into four main categories: color transformation, geometric transformation, context-based tasks, and cross-modal based tasks. These pretext tasks have been used in various scenarios based on the problem intended to be solved.

\subsection{Color Transformation}

\begin{figure}[ht]
  \centering
  \includegraphics[width=0.4\textwidth]{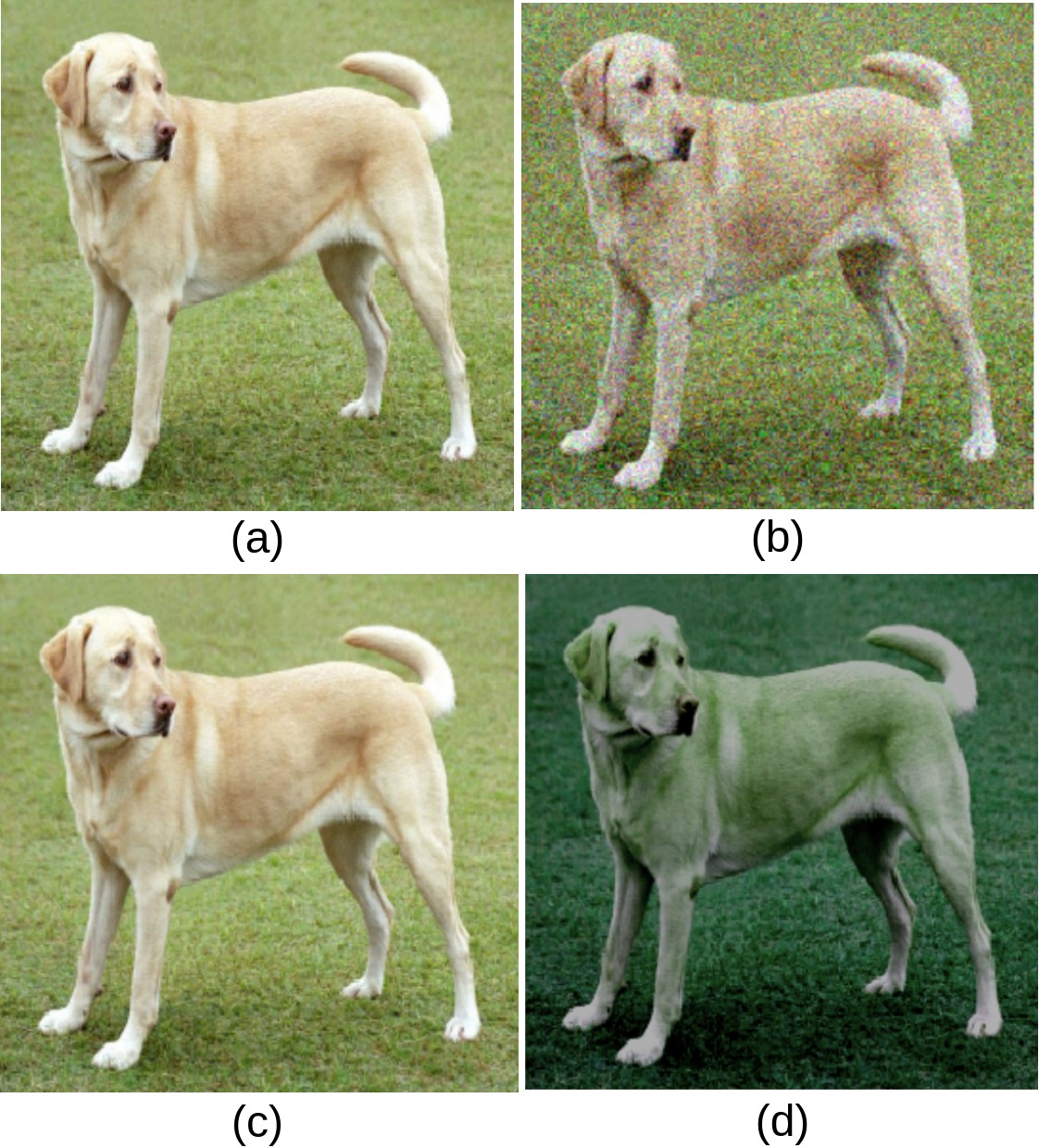}
  \caption{Color Transformation as pretext task \cite{chen2020simple}. (a) Original (b) Gaussian noise (c) Gaussian blur (d) Color distortion (Jitter) }\label{color_transformation}
\end{figure}

Color transformation involves basic adjustments of color levels in an image such as blurring, color distortions, converting to grayscale, etc. Figure \ref{color_transformation} represents an example of color transformation applied on a sample image from the ImageNet dataset \cite{chen2020simple}.  During this pretext task, the network learns to recognize similar images invariant to their colors.

\subsection{Geometric Transformation}

\begin{figure}[h!]
  \centering
  \includegraphics[width=0.4\textwidth]{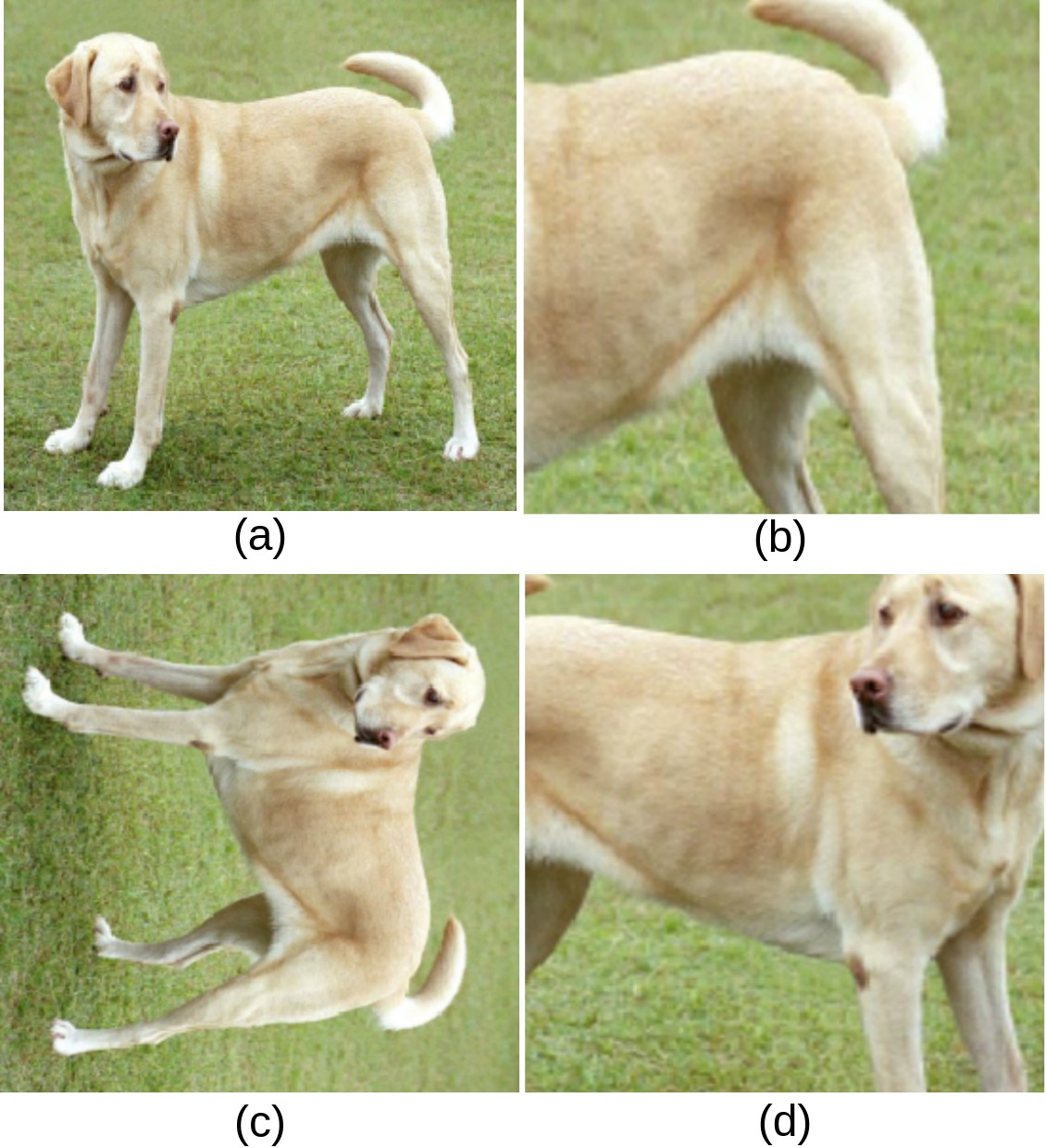}
  \caption{Geometric Transformation as pretext task \cite{chen2020simple}. (a) Original (b) Crop and Resize (c) Rotate(90$^\circ$, 180$^\circ$, 270$^\circ$) (d) crop, resize, flip}\label{geo-transformation}
\end{figure}

A geometric transformation is a spatial transformation where the geometry of the image is modified without altering its actual pixel information. The transformations include scaling, random cropping, flipping (horizontally, vertically), etc. as represented in figure \ref{geo-transformation} through which global-to-local view prediction is achieved.  Here the original image is considered as the global view and the transformed version is considered as the local view. Chen et. al. \cite{chen2020simple} performed such transformations to learn features during pretext task.

\subsection{Context-Based}

\subsubsection{Jigsaw puzzle}
\begin{figure}[ht]
  \centering
  \includegraphics[width=0.4\textwidth]{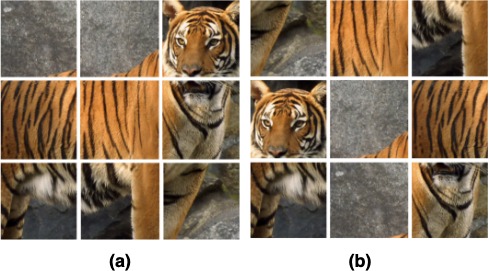}
  \caption{Solving jigsaw puzzle being used as a pretext task to learn representation.  (a) Original Image (b) reshuffled image.  The original image is the anchor and the reshuffled image is the positive sample.  }\label{jigsaw}
\end{figure}

Traditionally, solving jigsaw puzzles has been a prominent task in learning features from an image in an unsupervised way.  It involves identifying the correct position of the scrambled patches in an image by training an encoder (figure \ref{jigsaw}).  In terms of contrastive learning, the original image is the anchor, and an augmented image formed by scrambling the patches in the original image acts as a positive sample.  The rest of the images in the dataset/batch are considered to be negative samples \cite{misra2019selfsupervised}. 

\subsubsection{Frame order based} 
This approach applies to data that extends through time. An ideal application would be in the case of sensor data or a sequence of image frames (video). A video contains a sequence of semantically related frames. This implies that frames that are nearby with respect to time are closely related and the ones that are far away are less likely to be related.  Intuitively, the motive for using such an approach is, solving a pretext task that allows the model to learn useful visual representations while trying to recover the temporal coherence of a video. Here, a video with shuffled order in the sequence of its image frames acts as a positive sample while all other videos in the batch/dataset would be negative samples.  


Similarly, other possible approaches include randomly sampling two clips of the same length from a longer video or applying spatial augmentation for each video clip.  The goal is to use a contrastive loss to train the model such that clips taken from the same video are arranged closer whereas clips from different videos are pushed away in the embedding space.
In the work proposed by Qian et. al. \cite{qian2020spatiotemporal}, the framework contrasts the similarity between two positive samples to those of negative samples.  The positive pairs are two augmented clips from the same video.  As a result, it separates all encoded videos into non-overlapping regions such that an augmentation used in the training perturbs an encoded video only within a small region in the representation space.

\subsubsection{Future prediction}

\begin{figure}[h!]
  \centering
  \includegraphics[width=0.8\textwidth]{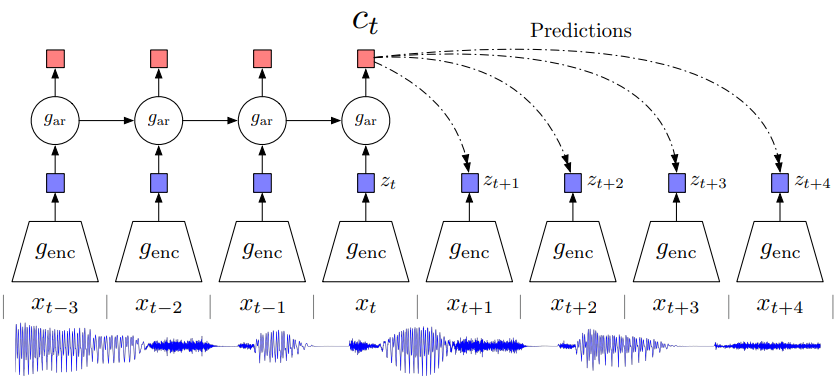}
  \caption{ Contrastive Predictive Coding:  Although the figure shows audio as input, similar setup can be used for videos, images, text etc. \cite{oord2018representation}}\label{cpc}
\end{figure}

One of the most common strategies for data that extends through time is to predict future or missing information.  This is commonly used for sequential data such as sensory data, audio signals, videos, etc. The goal of a future prediction task is to predict high-level information of future time-step given a series of past ones.  
In the work proposed by \cite{oord2018representation, lorre2020temporal},
high-dimensional data is compressed into a compact lower-dimensional latent embedding space.  Powerful autoregressive models are used to summarize the information in the latent space and a context latent representation $C_t$ is produced as represented in figure \ref{cpc}.  When predicting future information, the target (future) and context $C_t$ are encoded into a compact distributed vector representation in a way that maximally preserves the mutual information of the original signals.

\subsection{View Prediction (Cross modal-based)}

\begin{figure}[h!]
  \centering
  \includegraphics[width=0.45\textwidth]{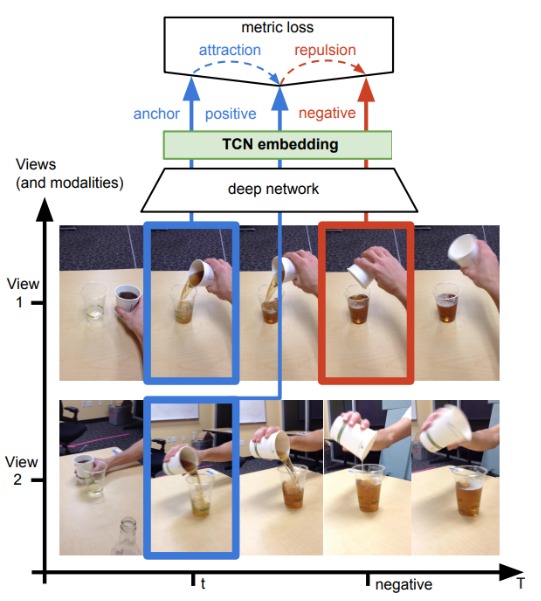}
  \caption{Learning representation from video frame sequence \cite{sermanet2017timecontrastive} }\label{time_contrastive}
\end{figure}

View prediction tasks are preferred for data that has multiple views of the same scene.  Following this approach, in \cite{sermanet2017timecontrastive}, the anchor and its positive images taken from simultaneous viewpoints, are encouraged to be close in the embedding space while distant from negative images taken from a different time within the same sequence.  The model learns by trying to simultaneously identify similar features between the frames from different angles and also trying to find the difference between frames that occur later in the sequence.  Figure \ref{time_contrastive} represents their approach for view prediction.  Similarly, recent work proposes an inter-intra contrastive framework where inter-sampling is learned through multi-view of the same sample, and intra-sampling that learns the temporal relation is performed through multiple approaches such as frame repetition and frame order shuffling that acts as the negative samples \cite{tao2020self}. 

\subsection{Identifying the right pre-text task}

\begin{figure}[h!]
  \centering
  \includegraphics[width=0.40\textwidth]{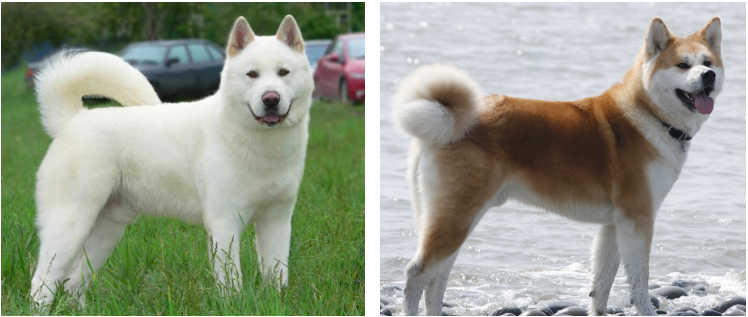}
  \caption{Most of the shapes of these two pairs of images are same.  However, low-level statistics are different (color and texture).  Usage of right pre-text task here is necessary\cite{noroozi2016unsupervised}}\label{fine_grain}
\end{figure}

The choice of pretext task relies on the type of problem being solved. Although numerous methods have been proposed in contrastive learning, a separate track of research is still going on to identify the right pre-text task.  Work has identified and proved that it is important to determine the right kind of pre-text task for a model to perform well with contrastive learning.  The main aim of a pre-text task is to compel the model to be invariant to these transformations while remaining discriminative to other data points. But the bias introduced through such augmentations could be a double-edged sword, as each augmentation encourages invariances to a transformation which can be beneficial in some cases and harmful in others.  For instance, applying rotation may help with view-independent aerial image recognition but might significantly downgrade the performance while trying to solve downstream tasks such as detecting which way is up in a photograph for a display application. \cite{xiao2020contrastive}.  Similarly, colorization-based pretext tasks might not work out in a fine-grain classification represented in figure \ref{fine_grain}.

\begin{figure}[h!]
  \centering
  \includegraphics[width=0.3\textwidth]{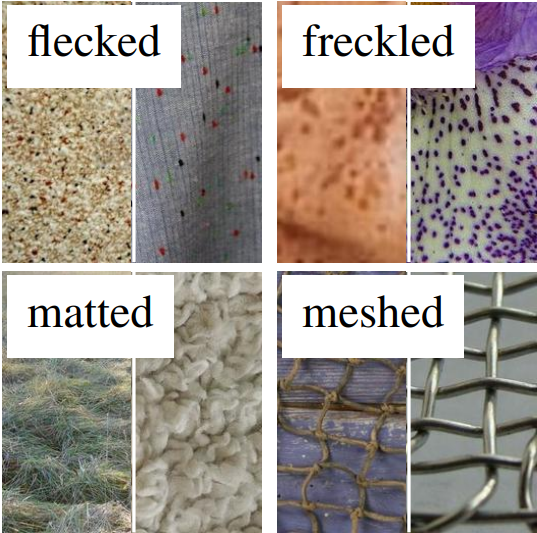}
  \caption{ A sample from the DTD dataset \cite{cimpoi2014describing}.  An example of why rotation based pretext task will not work well.}\label{rotation_issue}
\end{figure}

Similarly, in work \cite{yamaguchi2019multiple}, the authors focus on the importance of using the right pretext task.  The authors pointed out that in their scenario, except for rotation, other transformations such as scaling and changing aspect ratio may not be appropriate for the pretext task because they produce easily detectable visual artifacts. They also reveal that rotation does not work well when the image in a target dataset is constructed by color textures as in DTD dataset \cite{cimpoi2014describing} as shown in figure \ref{rotation_issue}.

\subsection{Pre-text tasks in NLP}
While self-supervised learning has been making significant progress in computer vision tasks for the past few years, it has been an active area of research in NLP for decades. Using a pretext task refers to generating labels such that supervised approaches can be applied to unsupervised problems to pre-train models. In NLP, text representations can be learned from large text corpora using any of the available pretext tasks that are discussed below.

\subsubsection{Center and Neighbor Word Prediction}
Back in 2013,  Word2Vec \cite{mikolov2013efficient} first introduced self-supervised methods to learn word representations in vector space. The continuous bag-of-words version of the model used "center word prediction" as the pretext task while the continuous skip-gram model implemented "neighbor word prediction" task. In center word prediction, the input to the model is a sequence of words with a fixed window size and one word missing from the center of the sequence. The task of the model is to predict the missing word in the sequence. On the other hand, the input in skip-gram model is a single word where the model predicts its neighbor words. By performing these particular tasks, the model is able to learn word representations that can be further used to train models for downstream tasks.

\subsubsection{Next and Neighbor Sentence Prediction}
In "next sentence prediction", the model predicts whether two inputs sentences can be consecutive sentences or not. A positive sample in this case would be a sample that follows the original sentence while a negative sample is a sentence from a random document. BERT \cite{devlin2018bert} used this method to drastically improve performance on downstream tasks that required an understanding of sentence relations such as question answering and language inference.

Similarly, given a sentence, a model has to predict its previous and the next sentence in "neighbor sentence prediction task". This approach was inherited by Skip-Thought Vectors \cite{kiros2015skip} paper. It is similar to the skip-gram method but rather applied to sentences in place of words.

\subsubsection{Auto-regressive Language Modeling}
This task involves predicting the next word, given previous words or vice-versa. A sequence of words from a text document is provided and the model tries to predict the next word that follows the sequence. This technique has been used by several n-gram models and neural networks such as GPT \cite{radford2018improving} and its recent versions.

\subsubsection{Sentence Permutation}
A recent paper known as BART \cite{lewis2019bart} used a pretext task where a continuous span of text from the corpus is taken and broken into multiple sentences. The position of the sentences are randomly reshuffled and the task of the model is to predict the original order of the sentences.
    
\section{Architectures}

\begin{figure*}[hbt!]
    \begin{subfigure}[b]{.24\textwidth}
        \centering
        \includegraphics[width=\textwidth]{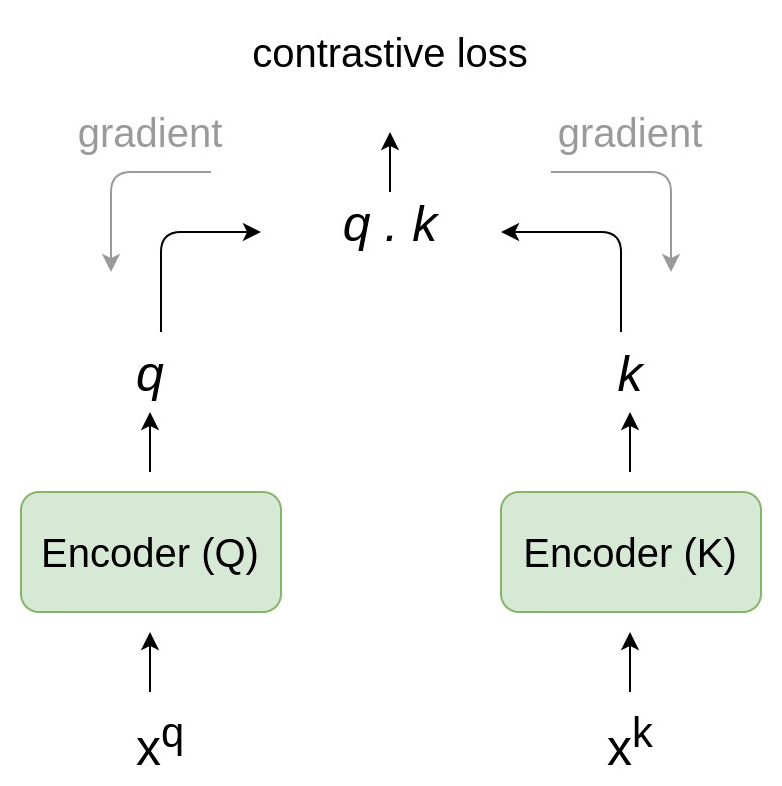}\hfill
        \caption{End-to-End}
        \label{fig:end_to_end}
    \end{subfigure}
    \begin{subfigure}[b]{.24\textwidth}
        \centering
        \includegraphics[width=\textwidth]{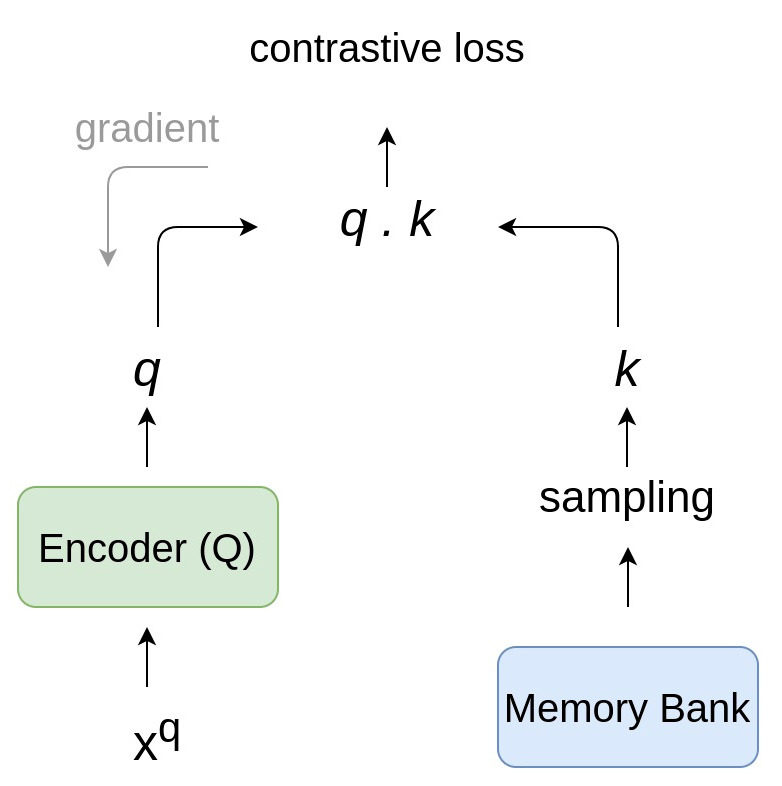}\hfill
        \caption{Memory Bank}
        \label{fig:memory_bank}
    \end{subfigure}
    \begin{subfigure}[b]{.24\textwidth}
        \centering
        \includegraphics[width=\textwidth]{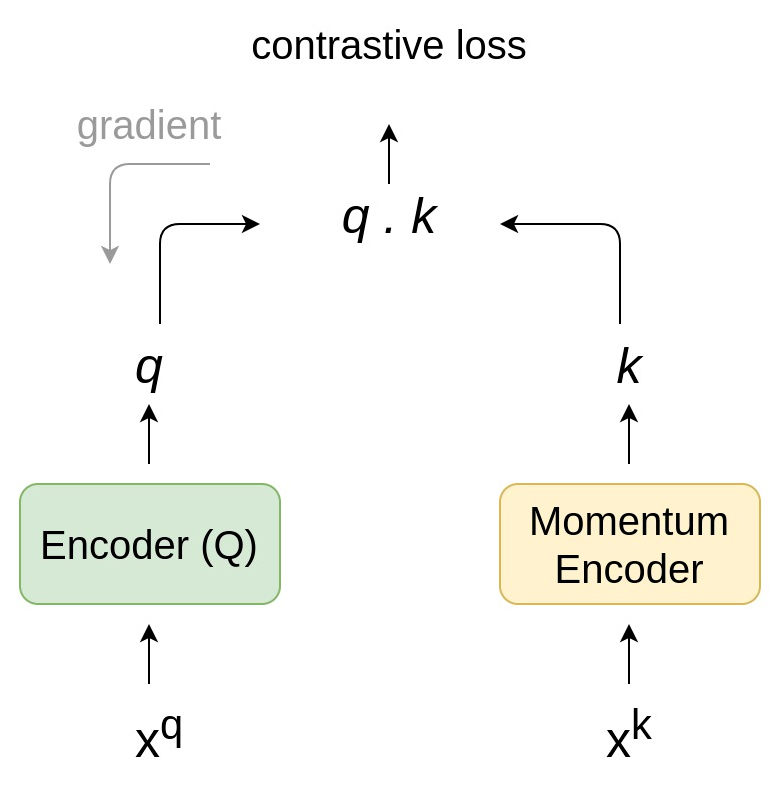}\hfill
        \caption{Momentum Encoder}
        \label{fig:momentum_encoder}
    \end{subfigure}
    \begin{subfigure}[b]{.24\textwidth}
        \centering
        \includegraphics[width=\textwidth]{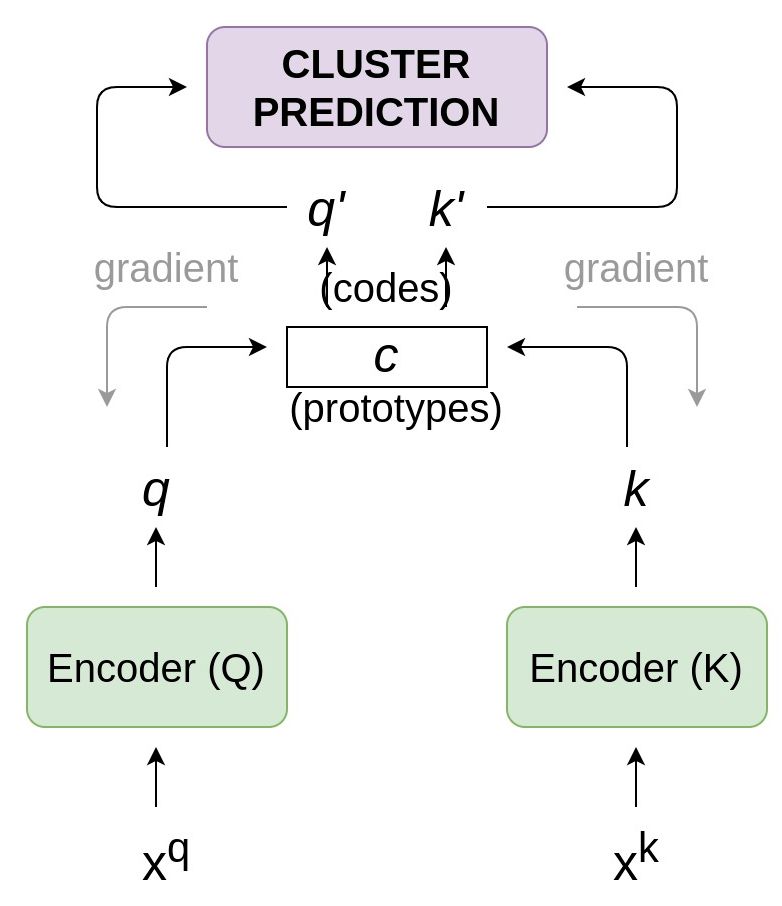}\hfill
        \caption{Clustering}
        \label{fig:clustering}
    \end{subfigure}
    \caption{Different architecture pipelines for Contrastive Learning: (a) End-to-End training of two encoders where one generates representation for positive samples and the other for negative samples (b) Using a memory bank to store and retrieve encodings of negative samples (c) Using a momentum encoder which acts as a dynamic dictionary lookup for encodings of negative samples during training (d) Implementing a clustering mechanism by using swapped prediction of the obtained representations from both the encoders using end-to-end architecture}
    \label{fig:architectures}
\end{figure*}

Contrastive learning methods rely on the number of negative samples for generating good quality representations. It can be seen as a dictionary-lookup task where the dictionary is sometimes the whole training set and the rest of the times some subset of the dataset. An interesting way to categorize these methods would be based on the technique used to collect negative samples against a positive data-point during training. Based on the approach taken, we categorized the methods into four major architectures as shown in figure \ref{fig:architectures}. Each architecture is explained separately along with examples of successful methods that follow similar principles.

\subsection{End-to-End Learning} \label{e2e}

End-to-end learning is a complex learning system that uses gradient-based learning and is designed in such a way that all modules are differentiable \cite{glasmachers2017limits}. This architecture prefers large batch sizes to accumulate a greater number of negative samples. Except for the original image and its augmented version, the rest of the images in the batch are considered negative. The pipeline employs two encoders: a Query encoder (Q) and a Key encoder (K) as shown in figure (\ref{fig:end_to_end}). The two encoders can be different and are updated end-to-end by backpropagation during training. The main idea behind training these encoders separately is to generate distinct representations of the same sample. Using a contrastive loss, it converges to make positive samples closer and negative samples far from the original sample. Here, the query encoder Q is trained on the original samples and the key encoder K is trained on their augmented versions (positive samples) along with the negative samples in the batch. The features q and k generated from these encoders are used to calculate the similarity between the respective inputs using a similarity metric (discussed later in section \ref{section:training}). Most of the time, the similarity metric used is "cosine similarity" which is simply the inner product of two vectors normalized to have length 1 as defined in equation \ref{eqn:similarity}.

\begin{figure}[h!]
  \centering
  \includegraphics[width=0.45\textwidth]{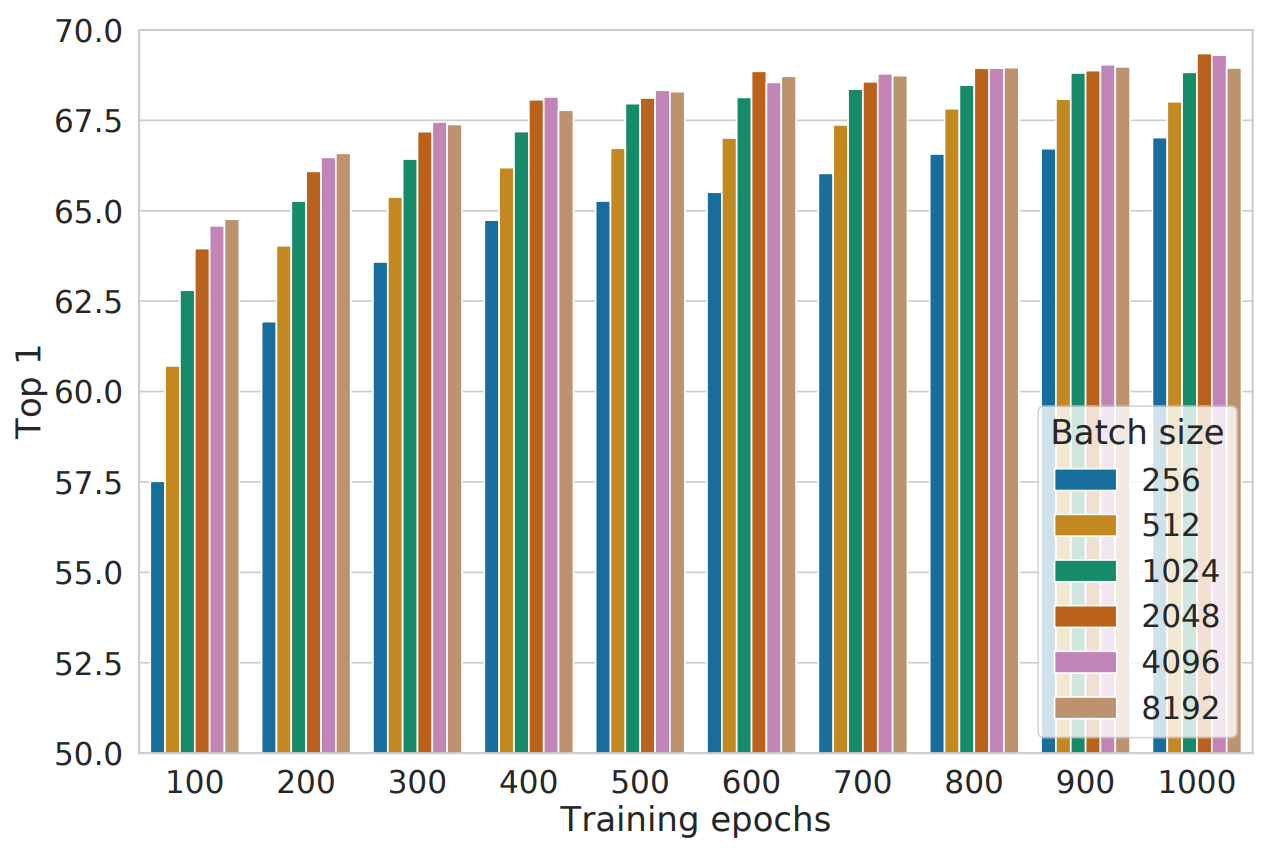}
  \caption{Linear evaluation models (ResNet-50) trained with different batch size and epochs.  Each bar represents a single run from scratch \cite{chen2020simple}}\label{endtoendperformance}
\end{figure}

Recently, a successful end-to-end model was proposed in SimCLR \cite{chen2020simple} where they used a batch size of 4096 for 100 epochs. It has been verified that end-to-end architectures are simple in complexity but perform better with large batch sizes and a higher number of epochs as represented in figure \ref{endtoendperformance}.  Another popular work that follows end-to-end architecture was proposed by Oord et. al \cite{oord2018representation} where they learn feature representations of high-dimensional time series data by predicting the future in latent space by using powerful autoregressive models along with a contrastive loss.  This approach makes the model tractable by using negative sampling.  Also, other works that follow this approach include \cite{hjelm2018learning, ye2019unsupervised, bachman2019learning, hnaff2019dataefficient, 20super}.  

The number of negative samples available in this approach is coupled with the batch size as it accumulates negative samples from the current batch. Since the batch size is limited by the GPU memory size, the scalability factor with these methods remains an issue. Furthermore, for larger batch sizes, the methods suffer from a large mini-batch optimization problem and require effective optimization strategies as pointed out by \cite{goyal2017accurate}.

\subsection{Using a Memory Bank} \label{memb}

With potential issues from having larger batch sizes that could inversely impact the optimization during training, a possible solution is to maintain a separate dictionary known as \textbf{memory bank}. 

\textbf{Memory Bank}:
The aim of maintaining a memory bank is to accumulate a large number of feature representations of samples that are used as negative samples during training. For this purpose, a dictionary is created that stores and updates the embeddings of samples with the most recent ones at regular intervals. The memory bank (M) contains a feature representation $m_I$ for each sample I in dataset D.  The representation $m_I$ is an exponential moving average of feature representations that were computed in prior epochs.  It enables replacing negative samples $m_{I^{'}}$ by their memory bank representations without increasing the training batch size.

\begin{figure}[hbt!]
\centering
    \includegraphics[width=0.50\linewidth]{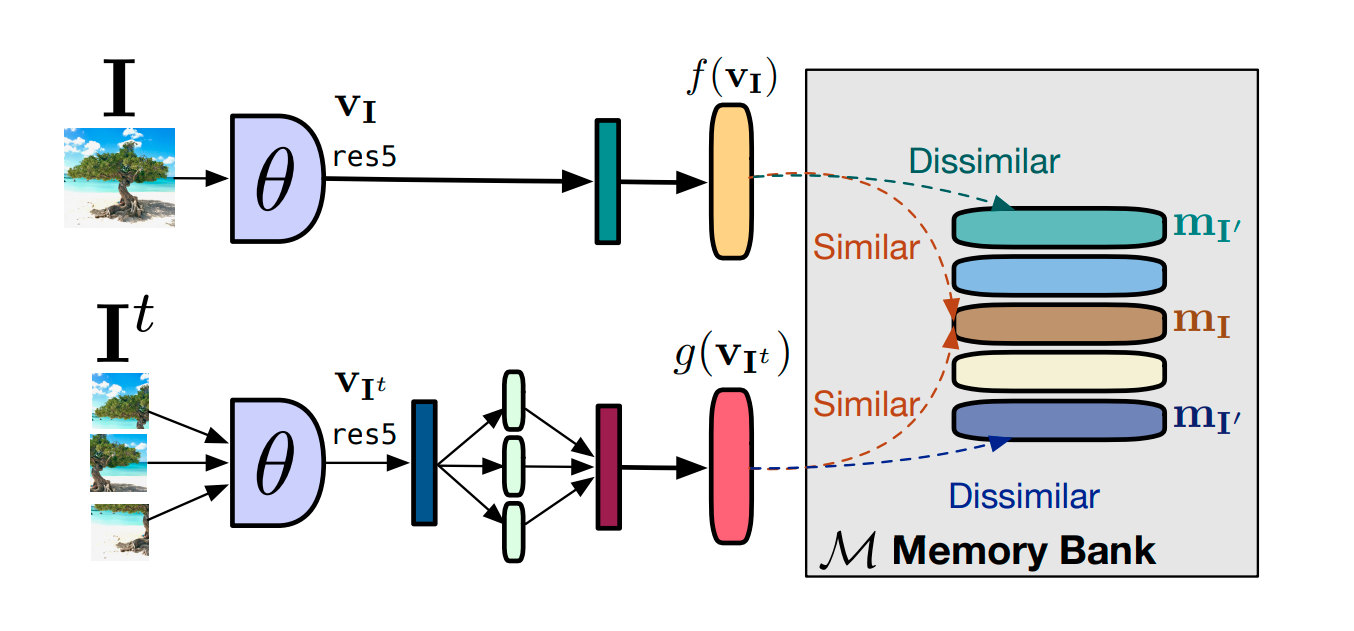}
    \caption{Usage of memory bank in PIRL: memory bank contains the moving average representations of all negative images to be used in contrastive learning. \cite{misra2019selfsupervised}}
    \label{fig:pirl_memory_bank}
\centering
\end{figure}

The representation of a sample in the memory bank gets updated when it is last seen, so the sampled keys are essentially about the encoders at multiple different steps all over the past epoch. PIRL \cite{misra2019selfsupervised} is one of the recent successful methods that learns good visual representations of images trained using a memory bank as shown in figure \ref{fig:pirl_memory_bank}. It requires the learner to construct representations of images that are covariant to any of the pretext tasks being used, though they focus mainly on the Jigsaw pretext task. Another popular work that uses a memory bank under contrastive setting was proposed by  Wu et al. \cite{wu2018unsupervised} where they implemented a non-parametric variant of softmax classifier that is more scalable for big data applications. 

However, maintaining a memory bank during training can be a complicated task. One of the potential drawbacks of this approach is that it can be computationally expensive to update the representations in the memory bank as the representations get outdated quickly in a few passes.

\subsection{Using a Momentum Encoder}

To address the issues with a memory bank explained in the previous section \ref{memb}, the memory bank gets replaced by a separate module called Momentum Encoder. The momentum encoder generates a dictionary as a queue of encoded keys with the current mini-batch enqueued and the oldest mini-batch dequeued.  The dictionary keys are defined on-the-fly by a set of data samples in the batch during training. The momentum encoder shares the same parameters as the encoder Q as shown in figure \ref{fig:momentum_encoder}.  It is not backpropagated after every pass, instead, it gets updated based on the parameters of the query encoder as represented by equation \ref{eqn:momentum_update} \cite{he2019momentum}.

\begin{equation}
    \theta_k \leftarrow	m\theta_k + (1 - m)\theta_q
    \label{eqn:momentum_update}
\end{equation}

In the equation, $m \in [0,1)$ is the momentum coefficient.  Only the parameters $\theta_q$ are updated by back-propagation.  The momentum update makes $\theta_k$ evolve smoothly than $\theta_q$.  As a result, though the keys in the queue are encoded by different encoders (in different mini-batches), the difference among these encoders can be made small.

The advantage of using this architecture over the first two is that it does not require training two separate models. Furthermore, there is no need  to maintain a memory bank that is computationally and memory inefficient.

\subsection{Clustering Feature Representations}
All three architectures explained above focus on comparing samples using a similarity metric and try to keep similar items closer and dissimilar items far from each other allowing the model to learn better representations. On the contrary, this architecture follows an end-to-end approach with two encoders that share parameters, but instead of using instance-based contrastive approach, they utilize a clustering algorithm to group similar features together.

\begin{figure}[hbt!]
\centering
    \includegraphics[width=0.85\linewidth]{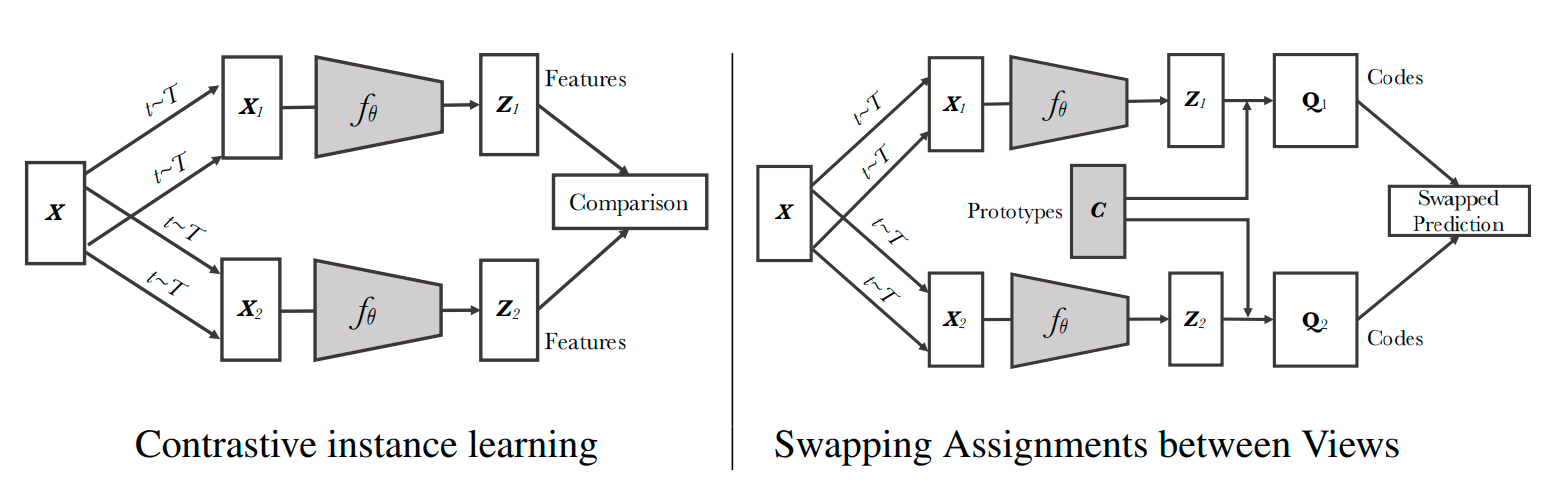}
    \caption{Conventional Contrastive Instance Learning v/s Contrastive Clustering of Feature Representations in SwAV \cite{caron2020unsupervised}}
    \label{fig:clustering_swav}
\centering
\end{figure}

One of the most recent works that employ clustering methods, SwAV \cite{caron2020unsupervised} is represented in figure \ref{fig:clustering_swav}.  The diagram points out the differences between other instance-based contrastive learning architectures and the clustering-based methods. Here, the goal is not only to make a pair of samples close to each other but also, make sure that all other features that are similar to each other form clusters together. For example, in an embedded space of images, the features of cats should be closer to the features of dogs (as both are animals) but should be far from the features of houses (as both are distinct).

In instance-based learning, every sample is treated as a discrete class in the dataset. This makes it unreliable in conditions where it compares an input sample against other samples from the same class that the original sample belongs to. To explain it clearly, imagine we have an image of a cat in the training batch that is the current input to the model. During this pass, all other images in the batch are considered as negative. The issue arises when there are images of other cats in the negative samples. This condition forces the model to learn two images of cats as not similar during training despite both being from the same class. This problem is implicitly addressed by a clustering-based approach.
    
\section{Encoders}
    \begin{figure}[hbt!]
\centering
    \includegraphics[width=0.50\linewidth]{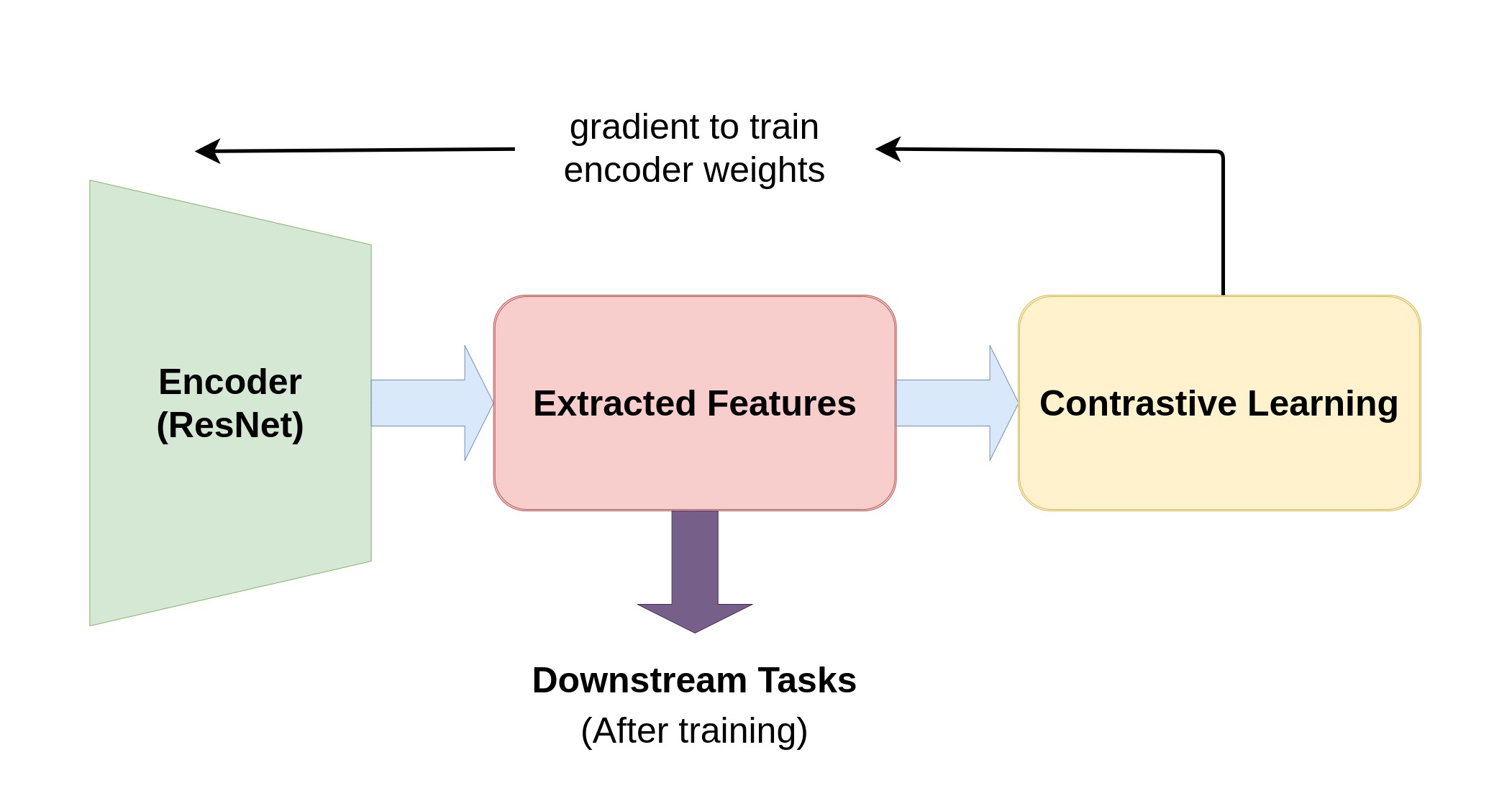}
    \caption{Training an Encoder and transfering knowledge for downstream tasks}
    \label{fig:encoder_training}
\centering
\end{figure}

Encoders play an integral role in any self-supervised learning pipeline as they are responsible for mapping the input samples to a latent space. Figure \ref{fig:encoder_training} reflects the role of an encoder in a self-supervised learning pipeline. Without effective feature representations, a classification model might have difficulty in learning to distinguish among different classes. Most of the works in contrastive learning utilize some variant of the ResNet \cite{he2016deep} model. Among its variants, ResNet-50 has been the most widely used because of its balance between size and learning capability.

 In an encoder, the output from a specific layer is pooled to get a single-dimensional feature vector for every sample. Depending on the approach, they are either upsampled or downsampled.  For example, in the work proposed by Misra et. al. \cite{misra2019selfsupervised}, a ResNet-50 architecture is used where the output of the res5 (residual block) features are average-pooled to get a 2048-dimensional vector for the given sample (image in their case).  They further apply a single linear projection to get a 128-dimensional feature vector.  Also, as part of their ablation test, they investigated features from various stages such as res2, res3, and res4 to evaluate the performance.  As expected, features extracted from the later stages of the encoder proved to be a better representation of the input than the features extracted from the earlier stages.
 
 Similarly, in the work proposed by Chen et. al. \cite{chen2019selfsupervised}, a traditional ResNet is used as an encoder where the features are extracted from the output of the average pooling layer.  Further, a shallow MLP (1 hidden layer) maps representations to a latent space where a contrastive loss is applied. For training a model for action recognition, the most common approach to extract features from a sequence of image frames is to use a 3D-ResNet as encoder \cite{lorre2020temporal, tao2020self}.

\section{Training} \label{section:training}
    To train an encoder, a pretext task is used that utilizes contrastive loss for backpropagation. The central idea in contrastive learning is to bring similar instances closer and push away dissimilar instances far from each other. One way to achieve this is to use a similarity metric that measures the closeness between the embeddings of two samples. In a contrastive setup, the most common similarity metric used is cosine similarity that acts as a basis for different contrastive loss functions. The cosine similarity of two variables (vectors) is the cosine of the angle between them and is defined as follows:

\begin{equation}
    cos\_sim(A, B) = \frac{A.B}{\|A\| \|B\|}
    \label{eqn:similarity}
\end{equation}

Contrastive learning focuses on comparing the embeddings with a Noise Contrastive Estimation (NCE) \cite{Gutmann2010NoisecontrastiveEA} function that is defined as follows:

\begin{equation}
    L_{NCE} = - log \frac{exp({sim(q, {k_+})}/\tau)}{exp({sim(q, {k_+})}/\tau) + exp({sim(q, {k_\_})}/\tau)}
    \label{eqn:NCE}
\end{equation}

where $q$ is the original sample, $k_+$ represents a positive sample, and $k_\_$ represents a negative sample. $\tau$ is a hyperparameter used in most of the recent methods and is called temperature coefficient. The sim() function can be any similarity function, but generally a cosine similarity as defined in equation \ref{eqn:similarity} is used. The initial idea behind NCE was to perform a non-linear logistic regression that discriminates between observed data and some artificially generated noise.

If the number of negative samples is greater, a variant of NCE called InfoNCE is used as represented in equation \ref{eqn:infoNCE}. The use of $L_{2}$ normalization (i.e. cosine similarity) and the temperature coefficient, effectively weighs different examples and can help the model learn from hard negatives.

\begin{equation}
    L_{infoNCE} = - log \frac{exp({sim(q, {k_+})}/\tau)}{exp({sim(q, {k_+})}/\tau) + \sum_{i=0}^{K} exp({sim(q, {k_i})}/\tau)}
    \label{eqn:infoNCE}
\end{equation}

where $k_i$ represents a negative sample.

Similar to other deep learning methods, contrastive learning employs a variety of optimization algorithms for training. The training process involves learning the parameters of encoder network by minimizing the loss function.

Stochastic Gradient Descent (SGD) has one of the most popular optimization algorithms used with contrastive learning methods \cite{misra2019selfsupervised, he2019momentum, bojanowski2017unsupervised,  wu2018unsupervised}. It is an stochastic approximation of gradient descent optimization since it replaces the actual gradient (calculated from the entire data set) by an estimate calculated from a randomly selected subset of data. A crucial hyperparameter for the SGD algorithm is the learning rate which in practice should gradually be decreased over time. An improved version of SGD (with momentum) is used in most deep learning approaches.

Another popular optimization method known as adaptive learning rate optimization algorithm (Adam) \cite{kingma2014adam} has been used in a few methods \cite{oord2018representation, srinivas2020curl, hafidi2020graphcl}. In Adam, momentum is incorporated directly as an estimate of the first-order moment. Furthermore, Adam includes bias corrections to the estimates of both the first-order moments and the second-order moments to account for their initialization at the origin.

Since some of the end-to-end methods \cite{chen2020simple, chen2020improved, caron2020unsupervised} use a very large batch size, training with standard SGD-based optimizers with a linear learning rate scaling becomes unstable. In order to stabilize the training, Layer-wise Adaptive Rate Scaling (LARS) \cite{you2017large} optimizer along with cosine learning rate \cite{loshchilov2016sgdr} was introduced. There are two main differences between LARS and other adaptive algorithms such as Adam. First, LARS uses a different learning rate for every layer that leads to better stability. Second, the magnitude of the update is based on the weight norm for better control of training speed. Furthermore, employing cosine learning rate involves periodically warm restarts of SGD, where in each restart, the learning rate is initialized to some value and is scheduled to decrease over time.


\begin{table*}
    \begin{center}
        \begin{tabular}{llrccc}
            \hline
            \multirow{2}{*}{\textbf{Method}} &   
            \multirow{2}{*}{\textbf{Architecture}} &    
              \multicolumn{2}{c}{\textbf{ImageNet (Self-supervised)}} &
              \multicolumn{2}{c}{\textbf{Semi-supervised (Top-5)}} \\
              & & Top-1 & Top-5 & 1\% Labels & 10\% Labels \\
            \hline
            Supervised & ResNet50 & 76.5 & - & 56.4 & 80.4 \\
            \hline
            CPC \cite{hnaff2019dataefficient} & ResNet v2 101 & 48.7 & 73.6 & - & - \\
            InstDisc \cite{wu2018unsupervised} & ResNet50 & 56.5 & - & 39.2 & 77.4 \\
            LA \cite{zhuang2019local} & ResNet50 & 60.2 & - & - & - \\
            MoCo \cite{he2019momentum} & ResNet50 & 60.6 & - & - & - \\
            BigBiGAN \cite{donahue2019large} & ResNet50 (4x) & 61.3 & 81.9 & 55.2 & 78.8 \\
            PCL \cite{li2020prototypical} & ResNet50 & 61.5 & - & 75.3 & 85.6 \\
            SeLa \cite{asano2019selflabelling} & ResNet50 & 61.5 & 84.0 & - & - \\
            PIRL \cite{misra2019selfsupervised} & ResNet50 & 63.6 & - & 57.2 & 83.8 \\
            CPCv2 \cite{hnaff2019dataefficient} & ResNet50 & 63.8 & 85.3 & 77.9 & \textbf{91.2} \\
            PCLv2 \cite{li2020prototypical} & ResNet50 & 67.6 & - & - & - \\
            SimCLR \cite{chen2020simple} & ResNet50 & 69.3 & 89.0 & 75.5 & 87.8 \\
            MoCov2 \cite{chen2020improved} & ResNet50 & 71.1 & - & - & - \\
            InfoMin Aug \cite{tian2020makes} & ResNet50 & 73.0 & 91.1 & - & - \\
            SwAV \cite{caron2020unsupervised} & ResNet50 & \textbf{75.3} & - & \textbf{78.5} & 89.9 \\
            \hline
        \end{tabular}
    
    \end{center}
    \caption{\textbf{Performance on ImageNet Dataset:} Top-1 and Top-5 accuracies of different contrastive learning methods on ImageNet using self-supervised approach where models are used as frozen encoders for a linear classifier. The second half of the table (rightmost two columns) show the performance (top-5 accuracy) of these methods after fine-tuning on 1\% and 10\% of labels from ImageNet}
    \label{table:imagenet}
\end{table*}


\section{Downstream Tasks}
    

\begin{figure}[hbt!]
\centering
    \includegraphics[width=0.55\linewidth]{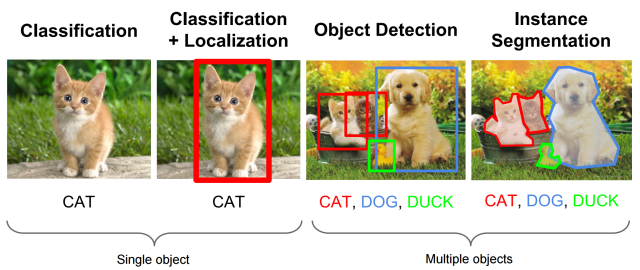}
    \caption{Image classification, localization, detection, and segmentation as downstream tasks in computer vision \cite{maj_2018}}
    \label{fig:downstream_vision}
\centering
\end{figure}

\begin{figure}[hbt!]
\centering
    \includegraphics[width=0.95\linewidth]{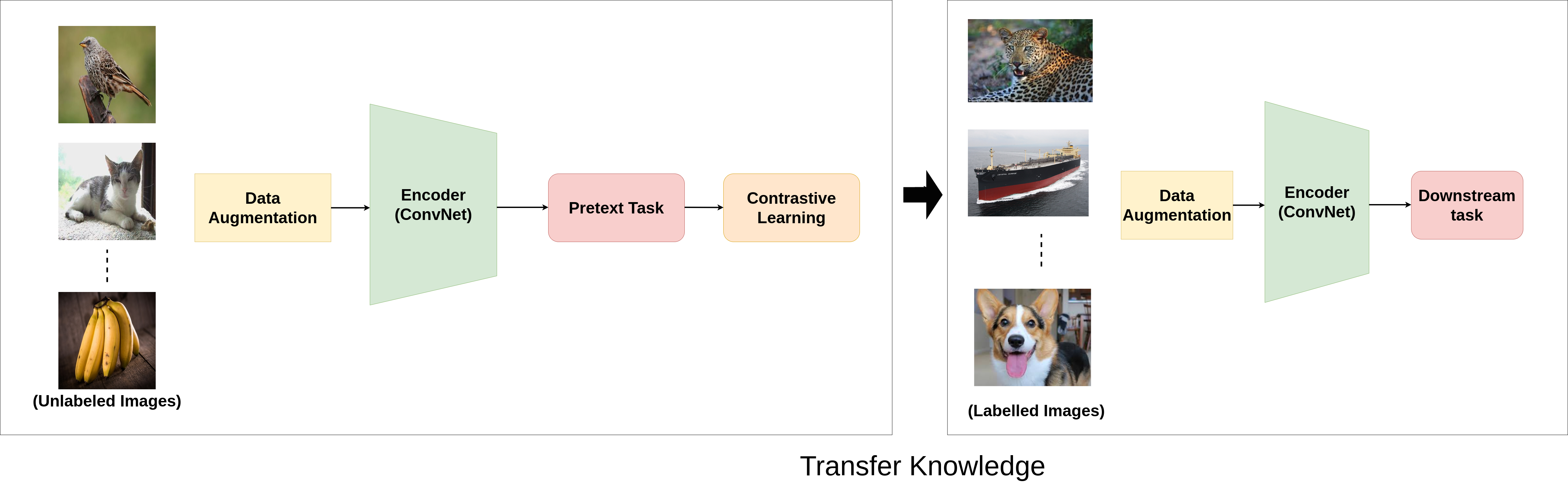}
    \caption{An overview of downstream task for images}
    \label{fig:downstream_for_images}
\centering
\end{figure}

Generally, computer vision pipelines that employ self-supervised learning involve performing two tasks: a pretext task and a downstream task.  Downstream tasks are application-specific tasks that utilize the knowledge that was learned during the pretext task.  They can be anything such as classification, detection, segmentation, future prediction, etc. in computer vision. Once example of downstream task can be hand gesture classification \cite{farahanipad2020hand} that involves both object detection and classification. Figure \ref{fig:downstream_for_images} represents the overview of how knowledge is transferred to a downstream task.  The learned parameters serve as a pretrained model and are transferred to other downstream computer vision tasks by fine-tuning.  The performance of transfer learning on these high-level vision tasks demonstrates the generalization ability of the learned features.

To evaluate the effectiveness of features learned with a self-supervised approach for downstream tasks, methods such as kernel visualization, feature map visualization, nearest-neighbor based approaches are commonly used to analyze the effectiveness of the pretext task.

\subsection{Visualizing Kernels and Feature Maps}

\begin{figure}[hbt!]
\centering
    \includegraphics[width=0.95\linewidth]{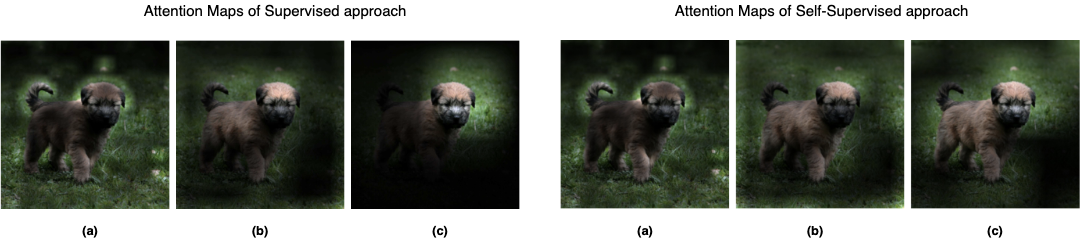}
    \caption{Attention maps generated by a trained AlexNet. The images represent the attention maps applied on features from Conv1 27x27, Conv3 13x13 and Conv5 6x6}
    \label{attnmaps}
\centering
\end{figure}

Here, the kernels of the first convolutional layer from encoders trained with both self-supervised (contrastive) and supervised approaches are compared. This helps to estimate the effectiveness of the self-supervised approach \cite{caron2019deep}. Similarly, attention maps generated from different layers of the encoders can be used to evaluate if an approach works or not. Gidaris et. al. \cite{gidaris2018unsupervised} assessed the effectiveness based on the activated regions observed in the input as shown in figure \ref{attnmaps}.

\subsection{Nearest Neighbor retrieval}
In general, the samples that belong to the same class are expected to be closer to each other in the latent space.  With the nearest neighbor approach, for a given input sample, top-K retrieval of the samples from the dataset can be used to analyze whether a self-supervised approach performs as expected or not.

\section{Benchmarks}

Recently, several self-supervised learning methods for computer vision tasks have been proposed that challenge the existing state-of-the-art supervised models. In this section, we collect and compare the performances of these methods based on the downstream tasks they were evaluated on. For image classification, two popular datasets ImageNet \cite{deng2009imagenet} and Places \cite{zhou2017places} have been used by most of the methods. Similarly, for object detection, Pascal VOC dataset has often been referred to for evaluation where these methods have outperformed the best supervised models. For action recognition and video classification, datasets such as UCF-101 \cite{soomro2012ucf101}, HMDB-51 \cite{kuehne2011hmdb}, and Kinetics \cite{carreira2017quo} have been used.

\begin{table*}
    \begin{center}
    \begin{tabular}{|l|c|c|c|}
        \hline
        \textbf{Method} & \textbf{Architecture} & \textbf{Parameters} & \textbf{Top-1 Accuracy}  \\
        \hline
        Supervised & ResNet50 & $25.6M$ & $53.2$\\
        \hline
        BiGAN \cite{donahue2017adversarial} & AlexNet & $61M$ & $31.0$\\
        Context \cite{doersch2016unsupervised} & AlexNet & $61M$ & $32.7$\\
        SplitBrain \cite{zhang2017splitbrain} & AlexNet & $61M$ & $34.1$\\
        AET \cite{zhang2019aet} & AlexNet & $61M$ & $37.1$\\
        DeepCluster \cite{caron2019deep} & AlexNet & $61M$ & $37.5$\\
        Color \cite{goyal2019scaling} & ResNet50 & $25.6M$ & $37.5$\\
        Jigsaw \cite{goyal2019scaling} & ResNet50 & $25.6M$ & $41.2$\\
        Rotation \cite{gidaris2018unsupervised} & ResNet50 & $25.6M$ & $41.4$\\
        NPID \cite{wu2018unsupervised} & ResNet50 & $25.6M$ & $45.5$\\
        PIRL \cite{misra2019selfsupervised} & ResNet50 & $25.6M$ & $49.8$\\
        LA \cite{zhuang2019local} & ResNet50 & $25.6M$ & $50.1$\\
        AMDIM \cite{bachman2019learning} & - & $670M$ & $55.1$\\
        SwAV \cite{caron2020unsupervised} & ResNet50 & $25.6M$ & \textbf{56.7}\\
        \hline
    \end{tabular}
    \end{center}
    \caption{Image classification accuracy on Places dataset pretrained on ImageNet.}
    \label{table:places}
\end{table*}

\begin{table*}
\begin{center}
\begin{tabular}{|l|c|c|c|c|}
\hline
\textbf{Method} & \textbf{Architecture} & \textbf{Parameters} & \textbf{(1)Classification} & \textbf{(2)Detection}  \\
\hline
Supervised & AlexNet & $61M$ & $79.9$ & $56.8$\\
Supervised & ResNet50 & $25.6M$ & $87.5$ & $81.3$\\
\hline
Inpaint \cite{pathak2016context} & AlexNet & $61M$ & $56.5$ & $44.5$\\
Color \cite{zhang2016colorful} & AlexNet & $61M$ & $65.6$ & $46.9$\\
BiGAN \cite{donahue2017adversarial} & AlexNet & $61M$ & $60.1$ & $46.9$\\
NAT \cite{bojanowski2017unsupervised} & AlexNet & $61M$ & $65.3$ & $49.4$\\
Context \cite{doersch2016unsupervised} & AlexNet & $61M$ & $65.3$ & $51.1$\\
DeepCluster \cite{caron2019deep} & AlexNet & $61M$ & $72.0$ & $55.4$\\
Color \cite{zhang2016colorful} & ResNet50 & $25.6M$ & $55.6$ & $-$\\
Rotation \cite{gidaris2018unsupervised} & ResNet50 & $25.6M$ & $63.9$ & $72.5$\\
Jigsaw \cite{goyal2019scaling} & ResNet50 & $25.6M$ & $64.5$ & $75.1$\\
LA \cite{zhuang2019local} & ResNet50 & $25.6M$ & $69.1$ & $-$\\
NPID \cite{wu2018unsupervised} & ResNet50 & $25.6M$ & $76.6$ & $79.1$\\
PIRL \cite{misra2019selfsupervised} & ResNet50 & $25.6M$ & $81.1$ & $80.7$\\
MoCo \cite{he2019momentum} & ResNet50 & $25.6M$ & $-$ & $81.4$\\
SwAV \cite{caron2020unsupervised} & ResNet50 & $25.6M$ & $88.9$ & $82.6$\\
\hline
\end{tabular}
\end{center}
\caption{(1) Linear classification top-1 accuracy on top of frozen features and (2) Object detection with finetuned features on VOC7+12 using Faster-CNN}
\label{table:pascal}
\end{table*}

Table \ref{table:imagenet} highlights the performance of several methods on ImageNet and reflects how these methods have evolved and performed better with time. At the moment, as seen in figure \ref{fig:cl_baselines}, SwAV \cite{caron2020unsupervised} produces comparable accuracy to the state-of-the-art supervised model in learning image representations from ImageNet.
Similarly, for image classification task on Places \cite{zhou2017places} dataset, SwAV \cite{caron2020unsupervised} and AMDIM \cite{bachman2019learning} have outperformed top supervised models with higher top-1 accuracies as shown in table \ref{table:pascal}. The methods shown in the table were first pretrained on ImageNet and later inferred on Places dataset using a linear classifier. The results advocate that representations learned by contrastive learning methods performed better than the supervised approach when tested on a different dataset.

These methods have not only excelled in image classification but also have performed well on other tasks like object detection and action recognition. As shown in table \ref{table:pascal}, SwAV \cite{caron2020unsupervised} outperforms the state-of-the-art supervised model in both linear classification and object detection in the Pascal VOC7 dataset. For linear classification, the models shown in the table were pretrained on VOC7 and features were taken for training a linear classification model. Similarly, for object detection, models were finetuned on VOC7+12 using Faster-RCNN. For video classification tasks, contrastive learning methods have shown promising results in datasets like UCF101, HMDB51, and Kinetics as reflected by table \ref{table:video_classification}.


\begin{table*}
\centering
\begin{tabular}{|l|c|c|c|c|c|}
\hline
\textbf{Method} & \textbf{Model} & \textbf{UCF-101} & \textbf{HMDB-51} &\textbf{K (top1)}  & \textbf{K (top5)} \\
\hline
C3D (Supervised) &-& 82.3 $^\dag$ & - & - & - \\
3DResNet-18 (Supervised) & R3D & 84.4$^\dag$ & 56.4$^\dag$ & - & - \\
P3D (Supervised) & - & 84.4$^\dag$ & - & - & - \\
\hline
ImageNet-inflated \cite{kim2019self} & R3D & 60.3 & 30.7 &-&- \\
jigsaw \cite{noroozi2016unsupervised} & - & 51.5 & 22.5 & - & - \\
OPN \cite{lee2017unsupervised}& - & 56.3 & 22.1 & - & - \\
Cross Learn (with Optical Flow)\cite{sayed2018cross} & - & 58.7 & 27.2&-&- \\
O3N \cite{fernando2017self} & - & 60.3 & 32.5 & -&- \\
Shuffle and Learn \cite{misra2016shuffle}& - & 50.2 & 18.1 & - & - \\
\hline
IIC (Shuffle + res)* \cite{tao2020selfsupervised} & R3D & 74.4 & 38.3 & - & - \\
inflated SIMCLR \cite{qian2020spatiotemporal} & R3D-50  & - & - & 48.0 & 71.5 \\
CVRL \cite{qian2020spatiotemporal} & R3D-50 & - & - & \textbf{64.1} & \textbf{85.8} \\
TCP \cite{lorre2020temporal} & R3D & 77.9 (3 splits) & 45.3 & - & - \\
SeCo inter+intra+order \cite{yao2020seco} & R3D & 88.26$^\dag$ & 55.5$^\dag$ & 61.91 & - \\
DTG-Net \cite{liu2020dtg} & R3D-18 & \textbf{85.6 }& \textbf{49.9} & - & - \\
CMC(3 views) \cite{tian2019contrastive} & R3D & 59.1 & 26.7 & - & - \\

\hline
\end{tabular}
\caption{Accuracy on Video Classification.  All the proposed methods were pre-trained with their proposed contrastive based approaches and a linear model was used for validation.  R3D in model represents 3D-ResNet.  $\dag$ represents that the model has been trained on another dataset and further fine-tuned with the specific dataset.  K represents Kinetics dataset}
\label{table:video_classification}
\end{table*}
    
\section{Contrastive Learning in NLP}
    Contrastive learning was first introduced by Mikolov et. al.\cite{mikolov2013distributed} for natural language processing in 2013. The authors proposed a contrastive learning-based framework by using co-occurring words as semantically similar points and negative sampling\cite{JMLR:v13:gutmann12a} for learning word embeddings. Negative sampling algorithm differentiates a word from the noise distribution using logistic regression and helps to simplify the training method. This framework results in huge improvement in the quality of representations of learned words and phrases in a computationally efficient way.
Arora et al.\cite{arora2019theoretical} proposed a theoretical framework for contrastive learning that learns useful feature representations from unlabeled data and introduced latent classes to formalize the notion of semantic similarity and performs well on classification tasks using the learned representations. Its performance is comparable to the state-of-the-art supervised approach on the Wiki-3029 dataset.
Another recent model, CONtrastive Position and Ordering with Negatives Objective(CONPONO) \cite{iter2020pretraining} discourses coherence and encodes fine-grained sentence ordering in text and outperforms BERT-Large model despite having the same number of parameters as BERT-Base.


Contrastive Learning has started gaining popularity on several NLP tasks in the recent years. It has shown significant improvement on NLP downstream tasks such as cross-lingual pre-training \cite{chi2020infoxlm}, language understanding \cite{fang2020cert}, and textual representations learning \cite{giorgi2020declutr}. INFOXLM \cite{chi2020infoxlm}, a cross-lingual pretraining model, proposes a  cross-lingual pretraining task based on maximizing the mutual information between two input sequences and learns to differentiate machine translation of input sequences using contrastive learning. Unlike TLM \cite{lample2019crosslingual}, this model aims to maximize mutual information between machine translation pairs in cross-lingual platform and improves the cross-lingual transferability in various downstream tasks, such as cross-lingual classification and question answering. Table \ref{nlp1} shows the recent contrastive learning methods on NLP downstream task.

\begin{table*}
\centering
\begin{tabular}{|l|l|l|}
\hline
Architecture & Dataset & Accuracy  \\
\hline\hline

INFOXLM \cite{chi2020infoxlm} & XNLI and MLQA & $79.7$ (AVG-5)\\
Distributed \cite{mikolov2013distributed}& Google internal & $72$\\
CERT \cite{fang2020cert} & QQP & $90.3$\\
CONPONO \cite{iter2020pretraining} & DiscoEval & $63.0$ (AVG-10)\\
Contrastive \cite{arora2019theoretical}& Wiki-3029 & $83.5$ (AVG-10)\\

\hline
\end{tabular}
\caption{Accuracy on different NLP dataset}
\label{nlp2}
\end{table*}

Most of the popular language models such as BERT \cite{devlin2018bert}, GPT \cite{radford2018improving} approach pretraining on tokens and hence may not capture sentence-level semantics. To address this issue, CERT \cite{fang2020cert} that pretrains models on the sentence level using contrastive learning was proposed. This model works in two steps: 1) creating augmentation of sentences using back-translation, and 2) predicting whether two augmented versions are from the same sentence or not by fine-tuning a pretrained language representation model (e.g., BERT, BART). CERT was also evaluated on 11 different natural language understanding tasks in the GLUE benchmark where it outperformed BERT on 7 tasks.
DeCLUTR \cite{giorgi2020declutr} is self-supervised model for learning universal sentence embeddings. This model outperforms InferSent, a popular sentence encoding method. It has been evaluated based on the quality of sentence embedding on the SentEval benchmark. Table \ref{nlp2} provides the comparison of accuracy on different NLP dataset.

\begin{table*}
\centering
\begin{tabular}{|l|l|l|}
    \hline
    \textbf{Model}  &  \textbf{Dataset} & \textbf{Application areas}\\
   
    \hline
    
    \hline

    Distributed Representations \cite{mikolov2013distributed} & Google internal & Training with Skip-gram model \\
    \hline

    Contrastive Unsupervised \cite{arora2019theoretical} & Wiki-3029 & Unsupervised representation learning
    \\
    \hline
    CONPONO \cite{iter2020pretraining} &	RTE, COPA, ReCoRD & Discourse fine-grained sentence ordering in text\\ 
    \hline
    INFOXLM \cite{chi2020infoxlm} &	XNLI and MLQA &	Learning cross-lingual representations\\
    \hline
    CERT \cite{fang2020cert}  &	GLUE benchmark &	Capturing sentence-level semantics\\

    \hline
    DeCLUTR \cite{giorgi2020declutr}	& OpenWebText &	Learning universal sentence representations\\
    \hline
\end{tabular}
\caption{Recent contrastive learning methods in NLP along with the datasets they were evaluated on and the respective downstream tasks}
\label{nlp1}
\end{table*}
    
\section{Discussions and Future Directions}

Although empirical results show that contrastive learning has decreased the gap in performance with supervised models, there is a need for more theoretical analysis to form a solid justification. For instance, a study by Purushwalkam et. al. \cite{purushwalkam2020demystifying} reveals that approaches like PIRL \cite{misra2019selfsupervised} and MoCo \cite{he2019momentum} fail to capture viewpoint and category instance invariance that are crucial components for object recognition. Some of these issues are further discussed below.



\subsection{Lack of Theoretical Foundation}

In an attempt to investigate the generalization ability of contrastive objective function, the empirical results from Arora et. al. \cite{arora2019theoretical} show that architecture design and sampling techniques also have a profound effect on the performance.  
Tsai et. al. \cite{tsai2020selfsupervised} provide an information-theoretical framework from a multi-view perspective to understand the properties that encourage successful self-supervised learning. They demonstrate that self-supervised learned representations can extract task-relevant information (with a potential loss) and discard task-irrelevant information (with a fixed gap). Ultimately, it propels the methods towards being highly dependent on the pretext task chosen during training. This affirms the need for more theoretical analysis on different modules in a contrastive pipeline.

\subsection{Selection of Data Augmentation and Pretext Tasks}

PIRL \cite{misra2019selfsupervised} emphasizes on methods that produce consistent results irrespective of the pretext task selected, but works like SimCLR \cite{chen2019selfsupervised}, MoCo-v2 \cite{chen2020improved} and Tian et. al. \cite{tian2020makes} demonstrate that selecting robust pretext tasks along with suitable data augmentations can highly boost the quality of the representations. Recently, SwAV \cite{caron2020unsupervised} beat other self-supervised methods by using multiple augmentations. It is difficult to directly compare these methods to choose specific tasks and transformations that can yield the best results on any dataset. 

\subsection{Proper Negative Sampling during Training}

During training, an original (positive) sample is compared against its negative counterparts that contribute towards a contrastive loss to train the model. In cases of easy negatives (where the similarity between the original sample and a negative sample is very low), the contribution towards the contrastive loss is minimal. This limits the ability of the model to converge quickly. To get more meaningful negative samples, top self-supervised methods either increase the batch sizes \cite{chen2020simple} or maintain a very large memory bank \cite{misra2019selfsupervised}. Recently, Kalantidis et. al. \cite{kalantidis2020hard} proposed a few hard negative mixing strategies to facilitate faster and better learning. However, this introduces a large number of hyperparameters that are specific to the training set and are difficult to generalize for other datasets.

\subsection{Dataset Biases}
In any self-supervised learning task, the data itself provides supervision. In effect, the representations learned using self-supervised objectives are influenced by the underlying data. Such biases are difficult to minimize with increase in the size of the datasets. 


    
\section{Conclusion}
    This paper has extensively reviewed recent top-performing self-supervised methods that follow contrastive learning for both vision and NLP tasks. We clearly explain different modules in a contrastive learning pipeline; from choosing the right pretext task, selecting an architectural design, to using the learned parameters for a downstream task. The works based on contrastive learning have shown promising results on several downstream tasks such as image/video classification, object detection, and other NLP tasks. Finally, this work concludes by discussing some of the open problems of current approaches that are yet to be addressed.  New techniques and paradigms are needed to tackle these issues.


\bibliographystyle{unsrt}  
\bibliography{references}  






\end{document}